\documentclass[10pt,conference,letter]{IEEEtran}
\IEEEoverridecommandlockouts

\usepackage{amssymb,amsmath,color,amsbsy, bm}

\usepackage{subcaption}
\usepackage{epsfig}
\usepackage{amsthm}

\newtheorem{question}{\textbf{Question}}

\usepackage{fixltx2e}
\usepackage{bbm}
\usepackage{ifthen}
\usepackage{verbatim}
\newboolean{showcomments}
\setboolean{showcomments}{true} 

\newcommand{\Meng}[1]{\ifthenelse{\boolean{showcomments}}
{ \textcolor{blue}{(Meng says:  #1)}}{}}

\newcommand{\shuo}[1]{\ifthenelse{\boolean{showcomments}}
{ \textcolor{blue}{(shuo explains:  #1)}}{}}
\usepackage{bm}
\usepackage{cite}

\usepackage{float}
\usepackage{xcolor}

\usepackage{graphicx}

\usepackage{amsmath,amsfonts}
 \newtheorem{theorem}{Theorem}

\newtheorem{proposition}{Proposition}
\newtheorem{lemma}{Lemma}
 \newtheorem{definition}{Definition}
 
\usepackage{algorithm}
\usepackage{algorithmic}

\begin{document}
%
\title{Minimizing Age of Information for Mobile Edge Computing Systems: A Nested Index Approach}

\author{Shuo Chen, 
        Ning Yang*, 
        Meng Zhang*, 
        Jun Wang

\thanks{Shuo Chen and Ning Yang are  with Institute of Automation, Chinese Academy of Sciences, Beijing, 100190, China. (e-mail: shuo.chen22@imperial.ac.uk, ning.yang@ia.ac.cn).

Meng Zhang is with the ZJU-UIUC Institute, Zhejiang University, Zhejiang, 314499, China. (e-mail: mengzhang@intl.zju.edu.cn).

Jun Wang is with the Department of Computer Science, University College London, WC1E 6BT, UK. (e-mail: jun.wang@cs.ucl.ac.uk).

({*}Corresponding author: Ning Yang, Meng Zhang)
}}
\maketitle

\begin{abstract}

Exploiting the computational heterogeneity of mobile devices and edge nodes, \textit{mobile edge computation} (MEC) provides an efficient approach to achieving real-time applications that are sensitive to information freshness, by offloading tasks from mobile devices to edge nodes. We use the metric \textit{Age-of-Information} (AoI) to evaluate information freshness. An efficient solution to minimize the AoI for the MEC system with multiple users is non-trivial to obtain due to the random computing time. In this paper, we consider multiple users offloading tasks to heterogeneous edge servers in a MEC system. We first reformulate the problem as a \textit{Restless Multi-Arm-Bandit} (RMAB) problem and establish a hierarchical \textit{Markov Decision Process} (MDP) to characterize the updating of AoI for the MEC system. Based on the hierarchical MDP, we propose a nested index framework and design a nested index policy with provably asymptotic optimality. Finally, the closed form of the nested index is obtained, which enables the performance tradeoffs between computation complexity and accuracy. Our algorithm leads to an optimality gap reduction of up to $40\%$, compared to benchmarks. Our algorithm asymptotically approximates the lower bound as the system scalar gets large enough.

\end{abstract}


%
\IEEEpeerreviewmaketitle

\section{Introduction}

\subsection{Motivation}
Large-scale cyber-physical applications necessitate real-time information. For example, Internet of Things (IoT) devices, constrained by limited computational resources, rely on cloud computing to boost performance, while sensor data from vehicles must be collected and processed to depict surroundings and facilitate navigation. Users demand prompt status updates. The \textit{Age-of-Information} (AoI) is a recently introduced metric designed to assess the freshness of information, quantifying the time elapsed since the most recent message update (e.g., \cite{yates_age_2020, Tripathi2021AoIGraphs}).

In numerous real-time applications, such as autonomous driving, the updated information is computationally demanding and necessitates processing. Offloading data to the cloud for computation can lead to data staleness and is computationally expensive. The \textit{Mobile Edge Computing} (MEC) paradigm shifts servers from the cloud to the edge, bringing users closer to servers and thereby reducing transmission delay (e.g., \cite{ mi2022MEC, tang2020deep, Ning2021MEC}). Consequently, MEC emerges as a promising technology capable of reducing latency and enhancing information freshness.

The majority of existing studies \cite{kadota_age_2021, Yates2019MultiSource, Kadota2018Unreliablechannels,tripathi_whittle_2019,liu_indexability_2008,hsu_age_2018,hsu_scheduling_2020,liu_delay-optimal_2016} primarily concentrate on optimizing AoI in MEC systems with a single user or server, or under the assumption of fixed computation time and task size. However, in practical scenarios, multiple heterogeneous users and servers are prevalent, prompting further exploration of heterogeneous MEC systems. Nevertheless, minimizing AoI in MEC systems with heterogeneous servers presents two challenges: determining the optimal location for task offloading and deciding the time for this offloading. To this end, we first answer the following question:

\begin{question}
How should one minimize the AoI in a MEC system with multiple heterogeneous users and servers?
\end{question}

The task of minimizing AoI is frequently formulated as a \textit{Restless Multi-Arm Bandit} (RMAB) problem, as it can be optimally solved by value iteration \cite{gittins2011multi}. However, such strategies are prone to the curse of dimensionality, necessitating near-optimal solutions with low complexity. A promising method for addressing the RMAB problem is the index policy approach  \cite{zou2021minimizing}, which is particularly suitable for scheduling systems with multiple nodes. This approach can yield near-optimal results with relatively low computational complexity. The effectiveness of the index policy is attributed to two primary factors: the ability to decompose the original problem into several sub-problems with practicable optimal solutions and the potential to express the index in closed form for a specific \textit{Markov Decision Process} (MDP) structure, thereby reducing computational complexity. Regrettably, neither of these factors can be easily assured: the optimal solution for the sub-problem may not exist, and obtaining the index function is non-trivial due to the presence of multiple state variables in MEC systems with heterogeneous users and servers. This leads us to the following question:

\begin{question}
How should we design an index-based policy for RMAB problems with multi-dimensional state variables?
\end{question}

\subsection{Solution Approach}
In response to this challenge, we suggest a framework where multiple heterogeneous users offload tasks to heterogeneous edge servers. We construct a multi-layer Markov Decision Process model aimed at minimizing the average AoI in MEC. The primary contributions of our research are as follows:

\begin{itemize} 
   \item[$\bullet$]\textit{Problem Formulation:}
    We formulate the problem of minimizing average AoI for MEC by optimizing offloading policies and reformulating it as an RMAB problem. \textit{To the best of our knowledge, this represents the first formulation of an age-minimal MEC problem that takes into account multiple heterogeneous users and edge servers.}
    
    \item[$\bullet$]\textit{Nested Index Approach:}
    We construct a multi-layer MDP model and, based on this, introduce a \textit{nested index framework} to solve our RMAB problem. We demonstrate the indexability of the multi-layer MDP for our MEC system and design the corresponding index function. We propose a nested index algorithm with provable asymptotic optimality.

    \item[$\bullet$]\textit{Numerical Results:}
    Our nested index algorithm results in an optimality gap reduction of up to $40\%$ compared to benchmarks. Our algorithm converges to the lower bound as the system scale increases sufficiently.
\end{itemize}

\section{Related Work}
\subsection{Age-of-Information}
Kaul \textit{et al.} in \cite{kaul2011minimizing} first proposed AoI as a metric to evaluate information freshness. The optimal AoI scheduling policy was to send messages from a source to the monitor through a single channel \cite{sun2017update}. In \cite{kadota_age_2021}, multiple sources could send updates over a single-hot network to a monitor, and they derived an approximate expression for the average AoI. In \cite{Yates2019MultiSource}, they minimized AoI by considering multiple sources for queuing systems. In \cite{Kadota2018Unreliablechannels}, they proposed a scheduling policy to minimize AoI in the wireless broadcast network with unreliable channels. In \cite{tripathi_whittle_2019}, they derived the structure of optimal policies for AoI minimizing problem and proved the optimality with reliable channel and unreliable channel assumptions. However, there was a lack of research on minimizing AoI in the more general MEC systems with heterogeneous multi-sources and edge servers.

\subsection{Restless Multi-Arm Bandit}

The RMAB problem arises when the state of an arm keeps changing whenever it is pulled or not \cite{whittle1988restless}. In \cite{liu_indexability_2008}, they formulated the problem of minimizing AoI as an RMAB problem and demonstrated that Whittle's Index was optimal when the arms were stochastically identical in a single-hop network. They also mentioned that a classic MDP was always indexable and proved the indexability of certain RMAB problems. Hsu \textit{et al.} \cite{hsu_age_2018} assumed that only one user could update at each time slot and obtained Whittle's Index in closed form. 
Hsu \textit{et al.} \cite{hsu_scheduling_2020} further studied the online and offline versions of the index approach and showed that the index policy was optimal when the arrival rate was constant. All the above index policies can only solve RMAB problems with one-dimensional state variables. However, in general, there exist more factors that affect decisions in wireless networks. Therefore, we need to consider multiple state variables for general wireless networks.

\subsection{Mobile Edge Computing}

In MEC scenarios, mobile edge servers are well equipped with sufficient computational resources and are close to users, enabling them to expedite the computation process. Yang \textit{et al.} \cite{8892492,9322150}  studied the resource management problem in MEC utilizing reinforcement learning approaches. In \cite{liu_delay-optimal_2016}, an MDP-based policy was proposed to determine whether to offload a task and when to transmit it. Zou and Ozel in \cite{zou2021optimizing} studied the transmission and computation process for MEC systems as coupled two queues in tandem. The computing time is random in the MEC system. The optimal scheduling policy contains non-preemptive \cite{sun2017update} and preemptive \cite{zou2021optimizing} structures, respectively. In \cite{zou2021optimizing}, the optimal scheduling policy under preemptive structure had a threshold property, and it was a benefit for minimizing AoI to wait before offloading. 
For minimizing AoI problems with multiple sources (or users), they established the MDP model and index-based policies \cite{zou2021minimizing,liu2010indexability}, which had less complex and relatively efficient. Such an index-based policy was proved to be asymptotic for many single-hop wireless network scheduling. To summarize, random offloading time and indeterminate computation durations considering preemptive techniques posed significant offloading challenges in the MEC system.

\section{System Model}
\subsection{System Overview}
\begin{figure}[t]
    \centering
    \includegraphics[width=0.28\textwidth]{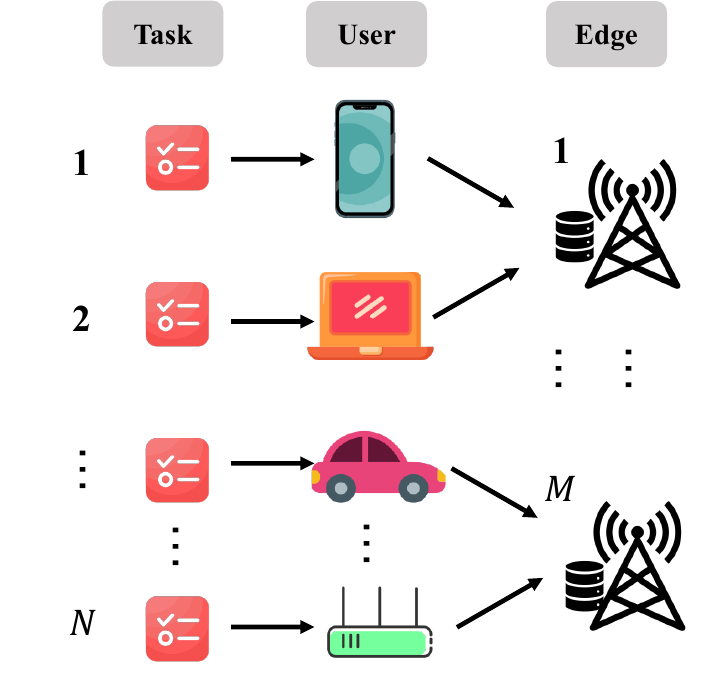}
    \caption{An MEC system in which each user offloads tasks to a certain edge server and receives data after processing.}
    \label{fig:sketch}
\end{figure}

We consider a MEC system with $N$ users who generate computational tasks and offload them to $M$ heterogeneous edge servers, as shown in Fig. \ref{fig:sketch}.
Let $n \in \mathcal{N},\ \mathcal{N}=\{1, 2, ..., N\}$ be the index of users and $m \in\mathcal{M},\ \mathcal{M}= \{1, 2, ..., M\}$ denote the index of edge servers. Let $t\in\mathcal{T}$ be the index of each time slot with $\mathcal{T} = \{ 1,2,...,T\} $. 
We consider the generate-at-will model \cite{sun2017update,hsu_age_2018}, i.e., the user can decide whether to generate a new task or not at each time slot $t$. The transmission time between users and edge servers is negligible, and once the computation of one task completes, its result is immediately sent back to the user. Each user can send a proportion of its task to any server for computing at each time slot \cite{partialoffloading}.

We assume that the edge servers are heterogeneous and the computing time of tasks is stochastic. Each task of user $n$ has a specific workload. When offloaded to a server, the task needs a specific number of CPU cycles to finish computing, and the computing time is based on both the workload and CPU frequency of the chosen server. We assume there is a minimum computing time $\tau_{n}^{min}$ for the task of user $n$.

We use the notion AoI to measure the freshness of information. We denote $\Delta_n(t)$ as the AoI for user $n$ at time $t$. 
The age of user $n$ decreases to the age of the latest offloaded task when the computing finishes or increases by $1$ otherwise.
Let $G_n(t)$ denote the generation time of the most recent task offloaded by user $n$ at time $t$. Then, the age of user $n$ at time $t$ if the computing finishes can be expressed as
\begin{equation}\label{gnt}
       \Delta_n(t) = t-G_n(t), \quad \forall n \in \mathcal{N}.
\end{equation}



\subsubsection{Offloading Decision} 
At time $t$, each user can choose one edge server to offload its tasks. When a task is offloaded, the computation starts at the beginning of each time slot. 
We denote $y_{nm}(t)\in\{0,1\}$ as the offloading decision variable for user $n$ at time $t$: $y_{nm}(t)=1$ if user $n$ decides to offload a task to server $m$. When $y_{nm}(t)=0,\forall m\in\mathcal{M}$, no task is to offload or the current task is to be dropped. 

Users' offloading decisions are subject to the following constraints:
\begin{subequations}\label{constraint}
    \begin{align}
    \sum\limits_{n\in\mathcal{N}}\sum\limits_{m\in\mathcal{M}}y_{nm}(t)\le M,&&\forall t\in\mathcal{T},\label{c1}\\
    \sum\limits_{m\in\mathcal{M}}y_{nm}(t)\le 1,&&\forall n\in\mathcal{N},t\in\mathcal{T},\label{c2}\\
    y_{nm}(t)\in \{1,0\},&&\forall n\in\mathcal{N},m\in\mathcal{M},t\in\mathcal{T}.\label{c3}
\end{align}
\end{subequations}
Specifically, constraint \eqref{c1} means there are at most $M$ servers to be chosen for offloading, and constraint \eqref{c2} means each user can offload its task to only one server at the same time. Constraint \eqref{c3} is a indicator function, which denotes whether a task is offload to server $m$ at time $t$.

\subsubsection{Shifted Geometric Distribution}\label{probability}
The transition of AoI during computation obeys a shifted geometric distribution \cite{taoexponential} with parameter $p_{m} = 1-e^{-\lambda_{m}}$ for tasks offloaded to server $m$, where $\lambda_{m}$ is the parameter of an exponential distribution.
We consider a minimal computational time for each computational task, denoted by $\tau^{min}_n$, i.e., only after $\tau^{min}_{n}$ time slots, 
each edge server $m$ completes the computation of the task with a probability $p_{m}$ within each time slot. Therefore, the transition probability of the AoI of each user $n$ during the computation can be written as
\begin{subequations}
\begin{align}\label{eq:tran_prob}
\mathbb{P}\{&\Delta_n(t+1) = \Delta_n(t) + 1  \mid \nonumber \\
& t-G_n(t)>\tau_n^{min},y_{nm}(t)=1\} = 1-p_{m}, \\
\mathbb{P}\{&\Delta_n(t+1) = t-G_n(t)+1\mid \nonumber\\
& t-G_n(t)>\tau_n^{min},y_{nm}(t)=1\} = p_{m},\\
\mathbb{P}\{&\Delta_n(t+1) = \Delta_n(t)+1\mid \nonumber\\
& t-G_n(t)\le\tau_n^{min},y_{nm}(t)=1\} = 1,\\
\mathbb{P}\{&\Delta_n(t+1) = \Delta_n(t)+1\mid y_{nm}(t)=0\} = 1.
\end{align}
\end{subequations}

\subsection{Problem Formulation}
We aim to minimize the overall AoI of the MEC system.
In the following, we formulate the AoI minimization problem.
Let $\pi\in\Pi$ denote the scheduling policy, which maps from the system state to the actions of all users.

We define the long-term average AoI \cite{yates_age_2020, hsu_scheduling_2020,zou2021minimizing} under policy $\pi$ as:
\begin{flalign}
\mathop{\lim\sup}\limits_{T\to\infty}\frac{1}{TN}\sum\limits_{t=1}^{T}\sum\limits_{n\in\mathcal{N}}\mathbb{E}_{\boldsymbol{y}\sim\pi}[\Delta_n(t)],
\end{flalign}
where we consider the policy $\pi$ is deterministic stationary \cite{sennott1989average}.

We reformulate the minimization problem of the long-term average AoI into the following form:
\begin{subequations}
\label{reformulatedproblem}
\begin{align}
\min\limits_{\pi}& \quad \mathop{\lim\sup}\limits_{T\to\infty}\frac{1}{TN}\sum\limits_{t=1}^{T}\sum\limits_{n\in\mathcal{N}}\mathbb{E}_{\boldsymbol{y}\sim\pi}[\Delta_n(t)]\\
    \mathrm{s.t.}&\quad \eqref{constraint} \nonumber
\end{align}
\end{subequations}
Based on the Lagrangian relaxation \cite{boyd2004convex}, we relax the instantaneous constraint (\ref{c1}) to average constraint, then drop constraint \eqref{c2} by introducing dual variables $\nu_m, \forall m\in\mathcal{M}$ and deriving $N$ sub-problems. Define $\pi_n\in\Pi_n$ as the policy which maps from the state of user $n$ to the action of user $n$. Given dual variables, each sub-problem $n$ is formulated as:
\begin{subequations}\label{originrelax}
\begin{flalign}
\ \ &\min\limits_{\pi_n}\ \ \mathop{\lim\sup}\limits_{T\to\infty}\frac{1}{TN} \!\!\sum\limits_{t=1}^{T}\mathbb{E}_{\bm{y}_n\sim\pi_n}\Big[\Delta_n(t)\!+\!\!\!\!\sum\limits_{m\in\mathcal{M}}\!\!\!\nu_m y_{nm}(t)\Big]\!&
\end{flalign}
\begin{align}\label{eq:decoupled_problem}
{\rm s.t.}\ &&\sum\limits_{m\in\mathcal{M}}y_{nm}(t)\le 1,&& \forall n\in\mathcal{N}, t\in\mathcal{T},\\
 &&y_{nm}(t)\in \{1,0\},&& \forall n\in\mathcal{N},m\in\mathcal{M}, \forall t\in\mathcal{T}.
\end{align}
\end{subequations}
When the dual variables converge, the sum of solutions to each sub-problem \eqref{originrelax} reaches the lower bound of the solution to problem \eqref{reformulatedproblem}, i.e., $\mathop{\lim\sup}\limits_{T\to\infty}\frac{1}{TN}\sum\limits_{t=1}^{T}\sum\limits_{n\in\mathcal{N}}\mathbb{E}_{\boldsymbol{y}\sim\pi}[\Delta_n(t)]\ge\sum\limits_{n\in\mathcal{N}}\mathop{\lim\sup}\limits_{T\to\infty}\frac{1}{TN} \!\!\sum\limits_{t=1}^{T}\mathbb{E}_{\bm{y}_n\sim\pi_n}\Big[\Delta_n(t)\!+\!\!\!\!\sum\limits_{m\in\mathcal{M}}\!\!\!\nu_m y_{nm}(t)\Big]$.
We will further study the optimal policy $\pi_n^*$ for the decomposed problem \eqref{originrelax} given $\{\nu_1,\nu_2,\dots,\nu_M\}$ by considering a multi-layer MDP.


\subsection{Multi-Layer MDP}

\begin{definition}[$L$-Layer MDP]\label{hmdp}
An $L$-layer MDP is a tuple $\langle\mathcal{S},\mathcal{A}, \mathcal{P}, \mathcal{C},L\rangle$, where $\mathcal{S}$ denotes the state space, $\mathcal{A}$ is the action space, the transition function is $\mathcal{P}:\mathcal{S}\times \mathcal{A} \to PD(\mathcal{S})$, the cost function is $\mathcal{C}:\mathcal{S}\times \mathcal{A} \to \mathbb{R}$, and $L\in\mathbb{Z}_+$ is the number of layers. Denote $\mathcal{S}_l=\mathbb{N}^l$ as the state space at layer $l$, and $\mathcal{S}=\mathcal{S}_1\cup\mathcal{S}_2\cup \cdots\cup\mathcal{S}_L$. An $L$-layer MDP fulfills the following conditions: $\mathcal{S}_l\subset\mathbb{N}^L$, and $\mathcal{S}=\mathcal{S}_1\cup\mathcal{S}_2\cup \cdots\cup\mathcal{S}_L$
\begin{itemize}
    \item  $\forall 0<l<L$ and $\forall s\in\mathcal{S}_l$, there exist some $a\in\mathcal{A}$ and $s'\in\mathcal{S}_{l+1}$ that satisfies $\mathbb{P}\{s' \mid s, a\}>0$,

    \item $\forall 0<l\le L$ and $\forall s\in\mathcal{S}_l$, there exist some $a\in\mathcal{A}$ and $s''\in\mathcal{S}_{l}$ that satisfies $\mathbb{P}\{s''\mid s, a\}>0$,

    \item $\forall L\ge l> 1$, there exists some $s\in\mathcal{S}_l$, $s'''\in\mathcal{S}_1$, and $a\in\mathcal{A}$ that satisfies $\mathbb{P}\{s'''\mid s, a\}>0$,
\end{itemize}
and we term the sub-space $\mathcal{S}_l$ as layer $l$.
\end{definition}

The multi-layer MDP defines the transition probability among states at different layers. The state at layer $l$ should be able to transit to states at layer $l$, $l+1$, and layer $1$. In a multi-layer MDP, states only transit among neighbor layers, which gives insights into the analysis of multi-dimensional state variables.


Now, we specify the multi-layer MDP for the MEC system.  Each sub-problem \eqref{originrelax} can be formulated as a $2$-layer MDP:
\begin{itemize}
    \item \textbf{Action space:} Let $\mathcal{A}=\{0,1\}^M$ be the action space for each user, and the action of user $n$, which is denoted as $\bm{y}_n(t)=\{y_{n1}(t),y_{n2}(t),\dots,y_{nM}(t)\}\in\mathcal{A}$, contains the information of offloading decisions. The action vector $\bm{y}_n(t)$ is composed entirely of zero elements, with at most one element being one.
    \item \textbf{State space:} Let $\mathcal{S}_l$ denote the state space for each user at layer $l$. Recall that we denote $\Delta_n(t)$ as the age of user $n$ and $G_n(t)$ denotes the generation time of the latest task of user $n$. User $n$ who is idle is defined at layer $1$, which has state $s_n(t)=\Delta_n(t)\in\mathcal{S}_1$. We consider users waiting for the result of the computation is at layer $2$. Let $D_n(t)=\Delta_n(G_n(t))$ denote the age of user $n$ when the latest task was generated. Thus we have state $s_n(t)=(\Delta_n(t),D_n(t))\in\mathcal{S}_2$.
    \item \textbf{Transition function:}
In the former section, we derive the transition probability in terms of $\Delta_n(t)$, while solely using $\Delta_n(t)$ can not fully characterize the transition of the states in multi-layer MDP. We use $q^{ss'}_{nm}$ to denote the transition probability of user $n$ from state $s$ to next state $s'$ by choosing server $m$ at time $t$, and we have
\begin{equation}
    q_{nm}^{ss'}=\mathbb{P}\{s'\ |\ s,\ y_{nm}(t)=1\},
\end{equation}
where $q_{nm}^{ss'}$ can be derived from the transition probability in Section \ref{probability}.
\item \textbf{Cost function:} We define the immediate cost as 
\begin{align}
    C_n(s_n(t),m)\triangleq \Delta_n(t) + \nu_m ,
\end{align}
which includes the current AoI and the server cost. 
\end{itemize}
According to \cite{sennott1989average}, it is simple to derive that there exists one deterministic stationary policy $\pi_n^*$ that reaches optimal average AoI. However, value iteration when deriving the optimal policy suffers from the curse of dimensionality \cite{optimalcontrol}. We need to seek an approach that owns less complexity and is near-optimal.

\section{Index-Based Policy}

In this section, we introduce a nested index approach to our RMAB problem, which is proven to be an asymptotically optimal offloading policy. First, we define the nested index and prove that the $2$-layer MDP for MEC systems fulfills the indexability condition. Next, we propose the nested index policy to schedule tasks. In addition, we also verify the asymptotic optimality of the proposed approach and obtain the nested index function in a closed form.

\subsection{Nested Index}
We first introduce the following definition of a \textit{passive set} based on \cite{whittle1988restless}. Define $\bm{\nu}\triangleq(\nu_1,\cdots,\nu_M)$ as the vector of \textit{activating cost}, where each $\nu_m$ is the server cost of choosing server $m$ for computation. We focus on sub-problem \eqref{originrelax} with given cost $\bm{\nu}$. We introduce the cost-to-go function which is a prediction of cost to evaluate the value of the state $s$. Denote the optimal average cost of sub-problem $n$ as $\gamma_n^*$, which is the minimum cost per stage. We can write the Bellman Equation of each sub-problem (\ref{eq:decoupled_problem}) as:
\begin{equation}
\begin{aligned}
        &\gamma_n^*+V_n(s, \bm{\nu})=\\
       & \min\limits_{m\in\mathcal{M},l\in\mathcal{L}}\left[ C_n(s,m)+\sum\limits_{s'\in\mathcal{S}_l}q^{ss'}_{nm}V_n(s',\bm{\nu})\right],
\end{aligned}\label{eq:BE}
\end{equation}
where function $V_n(s, \bm{\nu})$ is the differential cost-to-go \cite{optimalcontrol}. 

Let
\begin{equation}\label{bellman}
    \mu_{nm}(s,\bm{\nu})=C_n(s,m)+\sum\limits_{{s}'\in\mathcal{S}}q^{ss'}_{nm}V_n({s}',\bm{\nu})-\gamma^*_n
\end{equation}
denote the expected cost of choosing server $m$ given state $s$. 

As the decision process for our multi-layer MDP is different from Whittle's Index, which involves multi-dimensional state variables and multiple feasible actions, it motivates us to consider a multi-layer index structure.

\begin{definition}[Passive Set]
The passive set for user $n$ to transit to layer $l$ at server $m$ given activating cost $\bm{\nu}$ is denoted as:
\begin{equation}
\begin{aligned}
       &\mathcal{P}_{nm}^l(\bm{\nu})\triangleq \\
    &\left\{s\in\mathcal{S}_l\ |\min_{m'\in\mathcal{M}, m'\neq m }\mu_{nm'}(s,\bm{\nu})\leq \mu_{nm}(s,\bm{\nu})\right\}.
    \label{eq:passiveset}
\end{aligned}
\end{equation}
 We denote $\mathcal{P}_{nm}(\bm{\nu})\triangleq\cup_{l=1}^L\mathcal{P}_{nm}^l(\bm{\nu})$ as the overall passive set.
\end{definition}



The passive set refers to the set of states at layer $l$ that are sub-optimal for selecting server $m$ for computing with activating costs $\bm{\nu}$.

In classic RMAB problems \cite{hsu_age_2018, kriouile_global_2021}, activating cost $\nu$ is a scalar. Whittle \cite{whittle1988restless} stated that if the cardinality of the passive set increases monotonically from $0$ to $+\infty$ as activating cost $\nu$ increases from $0$ to $+\infty$, the problem is \textit{indexable}. Each state is assigned a maximum activating cost, which makes the same cost-to-go to take action or not at this state. The activating cost also gives the urgency of such a state as Whittle stated \cite{whittle1988restless}. A state with a higher activating cost has a higher priority for selection.

In MEC systems, we have multiple actions to choose from for each user, and we have heterogeneous servers with varying activating costs, which complicates the definition of \textit{indexability}. Therefore, we need to design a more sophisticated index technique.

\begin{definition}[Intra-Indexability]\label{Def4}
\textit{In a Multi-Layer MDP} $\langle{\mathcal{S},\mathcal{A}, \mathcal{P}, \mathcal{C},L}\rangle$, given servers cost $\bm{\nu}$, if for any layer $l$, the cardinality of passive set $|\mathcal{P}_{nm}^l(\bm{\nu})|$ increases monotonically to the cardinality $|\mathcal{S}_l|$ of layer $l$ as cost $\nu_m$ for server $m$ increases from $0$ to $+\infty$, then this multi-layer MDP is intra-indexable.
\end{definition}

The intra-indexability describes the relation between server costs and the optimal state to choose server $m$ at each layer. Given layer $l$, there exists the largest server cost $\nu_m'$ that state $s_n(t)$ is no longer included in the passive set $\mathcal{P}^l_{nm}(\bm{\nu})$ at layer $l$. The monotonic property guarantees the uniqueness of such an activating cost. The cardinality passive set over all layers $|\mathcal{P}_{nm}(\bm{\nu})|$ is non-decreasing in $\bm{\nu}$ if $|\mathcal{P}_{nm}^l(\bm{\nu})|$ is non-decreasing in $\bm{\nu},\forall 1\le l\le L$.

It is non-trivial to derive the optimal state at layer $l$ for server $m$ through the Bellman Equation Eq.\eqref{eq:BE}. We can, however, conclude the structure-property of the optimal solution for each sub-problem \eqref{originrelax}. As there are multiple layers in the multi-layer MDP, the optimal solution has a \textit{Multi-Layer-Threshold Type} (MLTT) structure:

\begin{definition}[Multi-Layer-Threshold Policy]Denote $\Tilde{m}=\mathop{\arg\max}\limits_{m\in\mathcal{M}} p_m$ as the index of the server that owns the best computational performance. If the following two conditions hold:
\begin{enumerate}
    \item If user $n$ is at layer 1 with state $s_n(t)=(\Delta_n(t)=A)$:
\begin{itemize}
    \item for any server $m'\neq \Tilde{m},m'\in\mathcal{M}$, there exists $H_n(m',\Tilde{m},0)$ that  $\forall A\ge \max\limits_{m'}\{H_n(m',\Tilde{m},0)\}$, $\pi_n(s_n(t))= \Tilde{m}$;
    \item for any two states $s_n(t), s_n(t')$ which fulfill $\Delta_n(t)<\Delta_n(t')$, then $p_{\pi_n(s(t))}\le p_{\pi_n(s(t'))}$;
\end{itemize}

\item If user $n$ is at layer 2 with state $s_n(t)=(\Delta_n(t)=A,D_n(t)=D)$:
\begin{itemize}
    \item given $D$, there exists $H_n(m',\Tilde{m},D)$ that the $\forall A\ge H_n(m,m',D)$, $\pi_n(s_n(t))= \Tilde{m}$;
    \item if $D_n(t)=D_n(t')$, $p_{\pi_n(s(t))}\le p_{\pi_n(s(t'))}$;
\end{itemize}
\end{enumerate}
then the offloading policy $\pi_n$ for the sub-problem 
\eqref{originrelax} $n$ has a Multi-Layer-Threshold Type (MLTT) structure.
\end{definition}

The MLTT structure shows some common properties for optimal thresholds at both layer $1$ and layer $2$. 
The proposition proposes that a threshold exists for a user when choosing between two servers, and the threshold depends only on the current age of the user given the same age at generation. Due to the limited space, all the
proofs are provided in the online appendix \cite{proof}.

\begin{proposition}\label{mltt}
The optimal solution $\pi_n^*$ to the sub-problem \eqref{originrelax} is MLTT.
\end{proposition}

The proof is shown in Appendix \ref{ap1}.

The MLTT is a more strict property than the threshold policy, as different actions have their own optimal thresholds on the age of users, and the threshold is determined by the layer $l$ and the generating age $D_n(t)$. By utilizing the MLTT property of the solution to problems in MEC systems, we can show the intra-indexability of the sub-problem \eqref{originrelax}.

\begin{theorem}\label{intraindex}
The MDP sub-problem \eqref{originrelax} is intra-indexable given cost $\bm{\nu}$.
\end{theorem}
The proof is shown in Appendix \ref{ap2}.

Since the AoI minimizing problem in MEC systems is a $2$-layer MDP, we design thresholds for actions at both layers. The intra-indexability property guarantees that there is one unique threshold for each server at each layer, and the size of the passive set for one server at each layer also increases as the server cost increases. Therefore, we can still use the index to represent the urgency of a state, and the comparison of states among layers is possible. We can then define the index for our $2$-layer MDP.






\begin{definition}[Nested Index]
\textit{Let $\bm{\nu}_{-m}$ denote the activating cost of edge servers except for server $m$. The nested index for taking $y_{n}(t)=m$ at state $s_n(t)$ is defined as}
\begin{equation}\label{nestedindex}
\begin{aligned}
 I_{nm}(s_n(t), \bm{\nu})&\triangleq&\\
\max\Big[0, \inf\Big\{\nu_m\ &|\ \min_{m'\in\mathcal{M}}\mu_{nm'}(s_n(t),[\bm{\nu}_{-m},\nu_m])\\
  &<\mu_{nm}(s_n(t),[\bm{\nu}_{-m},\nu_m])\Big\}\Big].
 \end{aligned}
\end{equation}
\end{definition}

\begin{figure}[!t]
\centering
\subfloat[The MC of adopting action $y_{nm}(t)=1$ at layer $1$, and the nested index is cost $\nu_m$ which makes $H_n(m,m',0)=d+2$.]{\includegraphics[width=2.2in]{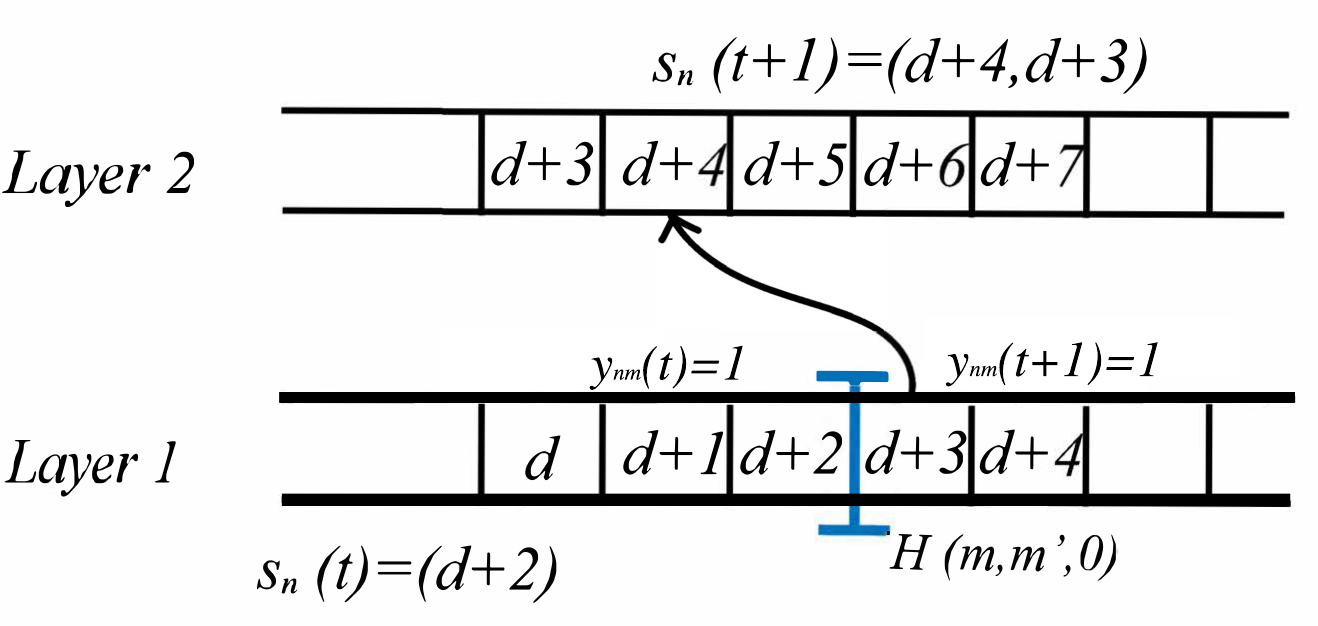}
\label{fig_first_case}}
\hfil
\subfloat[The MC of adopting action $y_{nm}(t)=1$ at layer $2$, and the nested index is cost $\nu_m$ which makes $H_n(m,m',d+3)=d+6$.]{\includegraphics[width=2.4in]{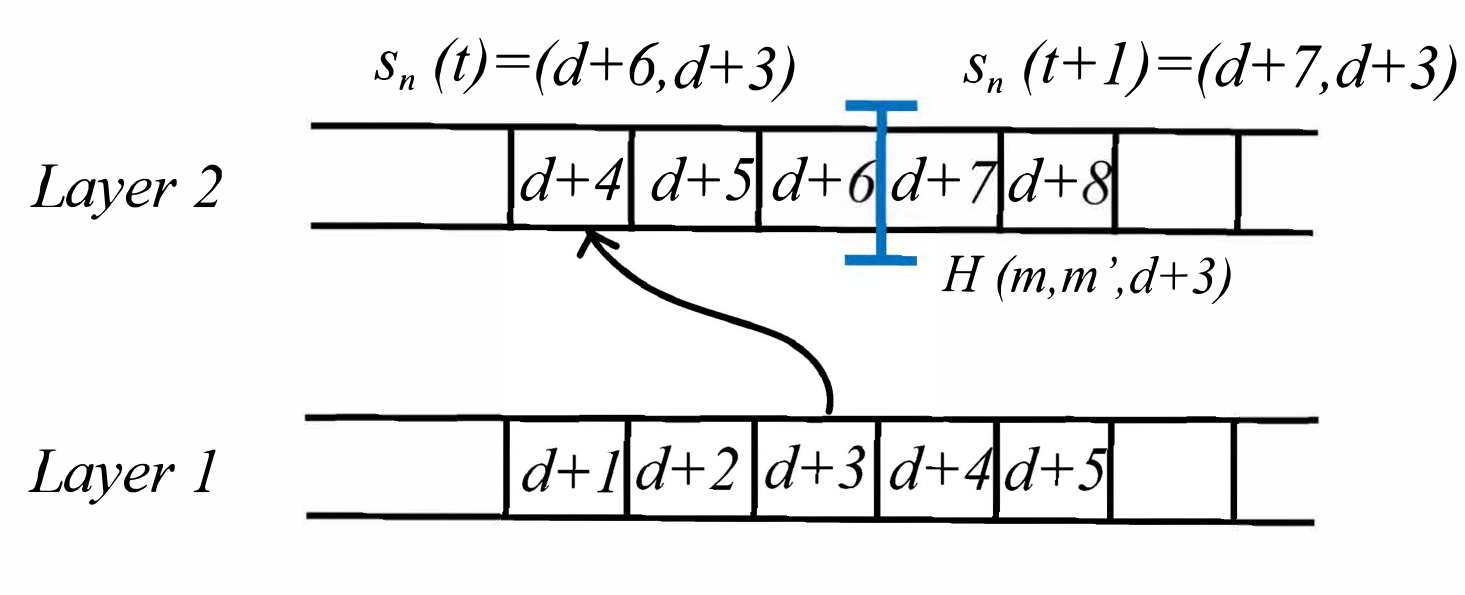}
\label{fig_2_case}}

\caption{The relationship between the \textit{nested-index} and the threshold in MC.}
\label{fig_mdpcost}
\end{figure}


The nested index allows us to characterize the urgency of each state. Compared with the partial index \cite{zou2021minimizing}, the nested index requires server costs other than server $m$, and the shows the emergency of transiting to different layers. 

Fig. \ref{fig_mdpcost} illustrates the relationship between the \textit{nested index} and the MDP. Fig. \ref{fig_first_case} compares the decision of offloading a task to server $m$ at neighbor time slots for user $n$ at layer 1. The optimal threshold $H_n(m,m',0)$ decreases as the server cost $\nu_m$ increases from 0 to $\infty$. The infimum of $\nu_m$ that makes $s_n(t)$, i.e., $\Delta_n(t)$, the optimal state to offload a task to server $m$ is the corresponding nested index. Fig. \ref{fig_2_case} illustrates the \textit{nested index} at layer 2. For simplicity, we consider user $n$ with the same generating age at time $t$ and $t+1$, i.e., $D_n(t)=D_n(t+1)$. The nested index can also be derived by adjusting $\nu_m$ when it is optimal to offload a task to server $m$ at time $t$. The \textit{nested index} gives the emergency of offloading tasks for a user at state $s_n(t)$.

\subsection{Nested Index Policy}
Based on the \textit{nested index} derived at each time slot, the central actuator can schedule the tasks of all users following the nested index policy. Define $u_{nm}$ as the decision variable for user $n$ on server $m$, and $w_{nm}=I_{nm}(s_n(t),\bm{\nu}(t))$ is the decision weight given by the nested index at time $t$.
We will solve the following binary decision scheduling problem at each time slot $t\in\mathcal{T}$:
\begin{subequations}\label{eq:algo}
    \begin{align}
         \label{algoobject1}\max\limits_{\bm{u}} &\sum\limits_{n\in\mathcal{N}}\sum\limits_{m\in\mathcal{M}}  I_{nm}(s_n(t),\bm{\nu}_{t-1})u_{nm}&\\\label{algocons1}
         s.t.&\sum\limits_{n\in\mathcal{N}}u_{nm}\le 1,&&\forall m\in\mathcal{M}, \\
         &\sum\limits_{m\in\mathcal{M}}u_{nm}\le 1,&&\forall n\in\mathcal{N},\\
         & u_{nm}\in \{0,1\} ,\forall n\in\mathcal{N},&&\forall m\in\mathcal{M}.
    \end{align}
\end{subequations}
We then make offloading decisions according to $y_{nm}(t)=u_{nm}$ at the solution. The decision variable $u_{nm}$ represents the policy that maps from the current state $s_n(t)$. The mapping process from the current state $s_n(t)$ to the action variable $y_{nm}(t)$ is named the \textit{nested index policy}.


\begin{algorithm}[t]

\caption{A Nested Index Policy}
    \begin{algorithmic}[1]\label{policy}
        \STATE Initialize parameters $N$, $M$, $L$, $\beta$;
        \STATE Initialize $s_n(0)$ for each user $n$ and server cost $\bm{\nu}_0$;
        \FOR{$t\le T$}
            \STATE Compute nested index $I_{nm}(s_n(t),\bm{\nu}_{t-1})$ for each user $n$ and layer $l$ by Eq. \eqref{nestedindex};
            \STATE Solve the maximization problem (\ref{eq:algo}) and obtains $u_{nm}$;
            \STATE Schedule tasks according to $y_{nm}(t)=u_{nm}$;
            \STATE Get state updates from edge servers and actions;
            \STATE Update cost $\bm{\nu}_{t}\gets(1-\beta)\bm{\nu}_{t-1}+\beta\bm{\nu}_{t-1}'$, where $\bm{\nu}_{t}$ is the server cost at time $t$, and $\bm{\nu}_{t}'$ is the optimal dual variable associate with problem \eqref{algocons1}.
        \ENDFOR 
    \end{algorithmic}
\end{algorithm}
In Algorithm \ref{policy}, we compute nested index $I_{nm}(s_n(t),\bm{\nu}_{t-1})$ for each user via Eq. (\ref{nestedindex}) and obtain the optimal solution $y_{nm}(t)$ for the problem (\ref{eq:algo}) which is a simple linear programming problem. Next, the solution for problem (\ref{eq:algo}) is mapped to offloading decisions and computing decisions to schedule tasks. We also get state updates from edge servers, offloading decisions, and computing decisions. Finally, we update activating cost $\bm{\nu}_{t}$. We execute lines 4-8 process at each time slot until the nested policy converges. 

The computation of the index value can be very complex, and we gave an approximation of the nested index given $s_n$, $\bm{\nu}_{-m}$, $\forall\  0<m<M$.

\begin{proposition}\label{indexfunc}
Given $s_n(t)=(\Delta_n(t),D_n(t))$, the index function satisfies
\begin{equation}
I_{nm}(s_n(t),\bm{\nu})=\nu_{m-1}+\Delta_n(t)-\gamma_n^*.
\end{equation}

The index for server $m$ can be derived by solving $I_{nm}(s_n(t),\bm{\nu})=\nu_m$.
 
\end{proposition}

The proof is given in Appendix \ref{ap3}. The index for other layers can be similarly derived within finite steps of computation. We derive the optimal average cost $\gamma_n^*$ by the technique similar to that used in \cite{tripathi_whittle_2019}, which involves solving a set of a finite number of equations. This reduces the complexity when computing the index function and makes our algorithm more feasible.

\subsection{Fluid Limit Model}

We use a fluid limit argument to show the optimality for the index policy in Algorithm 1 as in \cite{verloop_asymptotically_2016}.
The fixed point solution is the solution for the fluid limit model of the original problem. 
We will show that the fixed point of the fluid limit model for problem (\ref{eq:algo}) is equivalent to that of problem (\ref{originrelax}).

We define the fluid fixed point and the fluid limit model as follows. Let $z_{ns}\in [0,1]$ denote the fraction of user $n$ in state $s$, where $\sum_{s\in\mathcal{S}}z_{ns}=1$. Let  $x_{nm}^{s}\in[0,1]$ denote the fraction of user $n$ combined with state $s$ at server $m$ given by the optimal solution of the relaxed problem \eqref{originrelax}. Let $\bm{\nu}^*$ is the associated dual variable when it converges, and $(\bm{x}^*,\bm{z}^*,\bm{\nu}^*)$ represent the \textit{fluid fixed point} of the following \textit{fluid limit reformulation} of problem (\ref{originrelax}):
\begin{subequations}\label{fluidx}
\begin{align}
\mathop{\rm{min}}\limits_{\bm{x},\bm{z}}& \sum_{n\in \mathcal{N}}{\sum_{s\in\mathcal{S}}}\sum\limits_{m\in\mathcal{M}}z_{ns}C_{ns}x_{nm}^{s}&\\
\rm{\mathop{s.t.}} &\sum_{n\in \mathcal{N}}\sum_{s\in \mathcal{S}}z_{ns}x_{nm}^{s}\le 1,&& \forall m\in\mathcal{M},\\
&\sum\limits_{m\in\mathcal{M}}x_{nm}^{s}\le 1,&&\forall n\in\mathcal{N}, \forall s\in\mathcal{S},\\
&\sum\limits_{s\in\mathcal{S}}z_{ns}= 1,&&\forall n\in\mathcal{N},\\
&z_{ns},x_{nm}^s\in[0,1],\forall n\in\mathcal{N},&&\forall m\in\mathcal{M},\forall s\in\mathcal{S},\\
&\sum_{s'\in \mathcal{S}}z_{ns}\sum\limits_{m\in\mathcal{M}}x_{nm}^{s} q^{ss'}_{nm}=&&\sum_{s'\in \mathcal{S}}z_{ns'}\sum\limits_{m\in\mathcal{M}}x_{nm}^{s'}q^{s's}_{nm},\nonumber\\
&\forall n\in\mathcal{N},\forall s\in\mathcal{S}.&
\label{fluidtrans}
\end{align}
\end{subequations}
where (\ref{fluidtrans}) is a fluid balance constraint \cite{verloop_asymptotically_2016}.

We can similarly derive the \textit{fluid limit reformulation} problem (\ref{fluidy}) for the scheduling problem (\ref{eq:algo}). Denote $z_{nm}'$ as the fraction of user $n$ in state $s$, $v_{nm}^s$ as the fraction of user n assigned with server $m$ under the index policy, and $\bm{\nu}'^*$ as the dual variable for the relaxed problem of the scheduling problem \eqref{eq:algo} when it converges. Let $(\bm{v}^*,\bm{z}'^*,\bm{\nu}'^*)$ be the fixed point solution, and we have:
\begin{subequations}
\begin{align}
\mathop{\rm{max}}\limits_{\bm{v},\bm{z}' }&\sum_{n\in \mathcal{N}}{\sum_{s\in\mathcal{S} }}\sum_{m\in\mathcal{M}}z_{ns}'w_{nm}^{s}v_{nm}^{s}\\
\label{fluidc1}
{\rm s.t.} &\sum_{n\in \mathcal{N}}\sum_{s\in \mathcal{S}}z_{ns}'v_{nm}^{s}\le 1, &&\forall m\in\mathcal{M},\\
&\sum_{m\in\mathcal{M}}v_{nm}^{s}\le 1,&& \forall n\in\mathcal{N},\forall s\in\mathcal{S}.
\end{align}\label{fluidy}
\end{subequations}
Then, we can evaluate the performance of our \textit{nested index} policy based on the fluid limit model for both problems.

\begin{proposition}\label{fixedpoint}
The fixed point solution to problem (\ref{fluidx}) is equivalent to the solution (the fluid fixed point) to problem (\ref{fluidy}), i.e., we have
\begin{equation}
    (\bm{v}^*,\bm{z}'^*,\bm{\nu}'^*)=(\bm{x}^*,\bm{z}^*,\bm{\nu}^*).
\end{equation}
\end{proposition}
The proof is shown in Appendix \ref{ap4}. The equivalence of fixed point solution builds the connection between problem (\ref{eq:algo}) and problem (\ref{eq:decoupled_problem}). Though problem (\ref{eq:algo}) follows an instantaneous constraint (\ref{algocons1}), its fixed point solution still reaches the optimality at the fluid limit, which contributes to the \textit{asymptotic optimality} of our policy in Algorithm \ref{policy}.

By scaling a system by $r$, we scale the number of users $N^r$ and servers $M^r$ by $r$ proportionally, i.e., let $N^r=r\cdot N$, $M^r=r\cdot M$ while keeping $\frac{N^r}{M^r}$ a constant.\footnote{The system parameters $\bm{\tau}$ and $\bm{p}$ are also scaled proportionally, i.e., $\bm{\tau}^r=[\bm{\tau},\bm{\tau},\dots,\bm{\tau}]\in\mathbb{Z}_+^{1\times N^r}$ and $\bm{p}^r=[\bm{p}',\bm{p}',\dots,\bm{p}']\in[0,1]^{M^r\times N^r}$, where $\bm{p}'=[\bm{p}^T,\bm{p}^T,\dots,\bm{p}^T]^T\in[0,1]^{M^r\times N}$.}
\begin{lemma}
Under a mild global attractor assumption,
the expected objective $V^{\pi}_r$ for problem (\ref{eq:algo})  under the nested index policy $\pi$ achieves the optimal objective $V^*$ for the fluid limit model of problem (\ref{eq:decoupled_problem}) asymptotically, i.e.,
\begin{equation}
\lim\limits_{r\to+\infty} V^{\pi}_r=V^*.
    \end{equation}
\end{lemma}

We refer to \cite{verloop_asymptotically_2016} for the details of the global attractor assumption. The equivalence of the fixed point solution shows the accordance of the fluid limit model for both problems. Under a mild global attractor assumption, the objective of problem \eqref{reformulatedproblem} under our \textit{nested index} converges to the optimal cost of problem \eqref{fluidx}.



\section{Numerical Results}

In this section, we perform numerical studies to evaluate the performance of our nested index algorithm and verify the convergence property. We simulate a MEC system with initial tasks of $N=50$ users that can be divided into $6$ groups, with 
$\bm{\tau}^{min}=[2, 4, 8, 16, 32, 64]$, and successful updating probability of $\bm{p}=[0.8, 0.7, 0.6, 0.5, 0.3, 0.1]$, each of which has the number of $[5,10,5,5,10,15]$ users respectively. We set $\beta=50$ and simulated $T=10000$ slots. In the simulation, we mainly test the performance of the \textit{nested index} when solving a multi-layer MDP. 
\begin{figure}[t]
    \centering
    \includegraphics[width=0.27\textwidth]{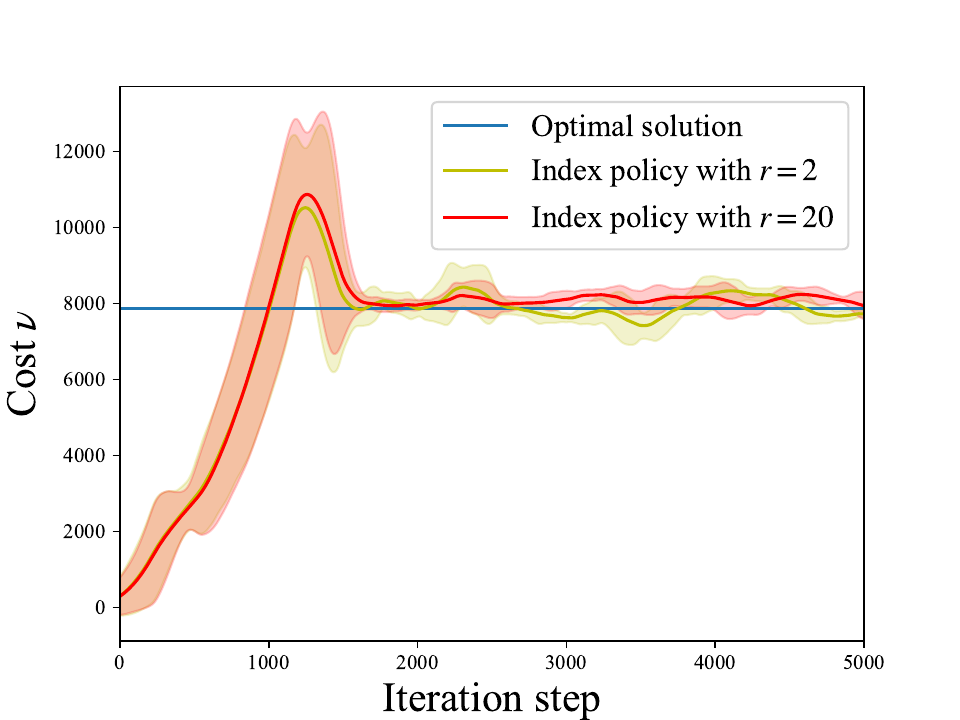}
    \caption{Dual cost update for the proposed index-based policy, which is compared with the dual cost update of the relaxed problem.}
    \label{fig:converge_lambda}
\end{figure}


\begin{figure}[t]
    \centering
    \includegraphics[width=0.27\textwidth]{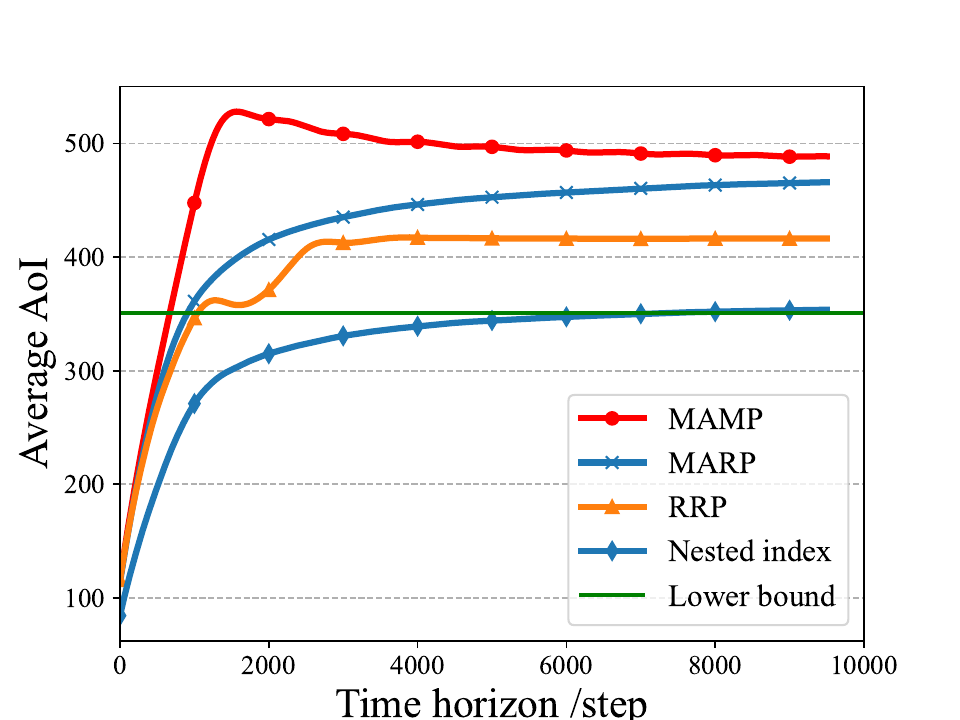}
    \caption{Average AoI performance during computing.}
    \label{aoi_converge}
\end{figure}
\begin{figure}[t]
\centering
\includegraphics[width=0.27\textwidth]{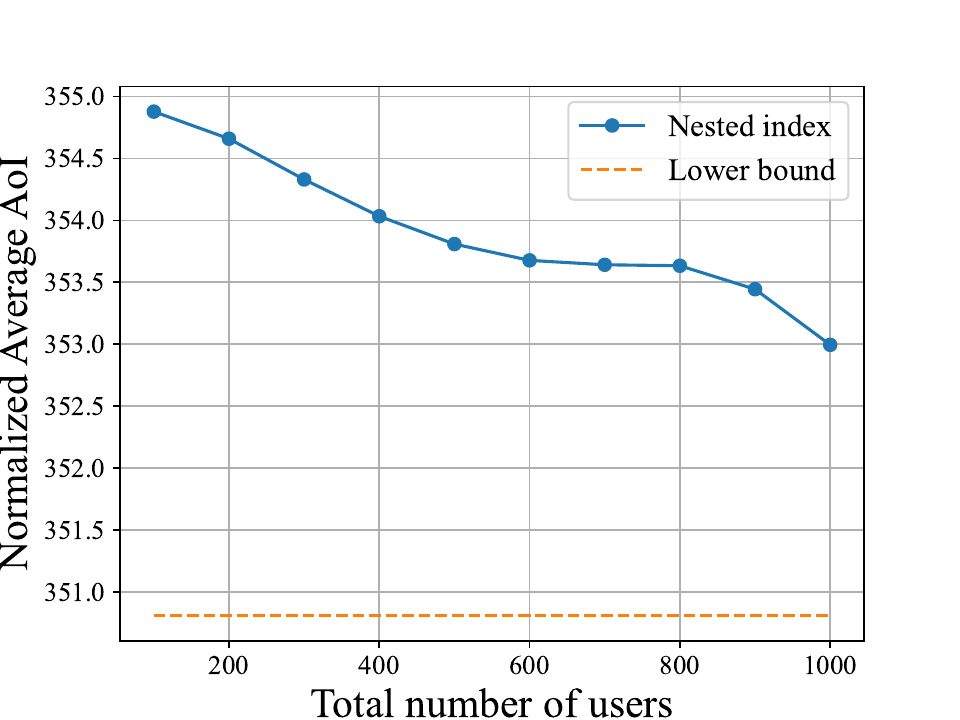}
\caption{Average AoI vs. system scale.}
\label{fig_sim}
\end{figure}
\subsection{Convergence of the Cost Update}
We compare the dynamics of the cost updates of the proposed nested index policy with that of the optimal solution to the problem (\ref{originrelax}). The optimal solution in Fig. \ref{fig:converge_lambda} represents the dynamics of the server cost $\nu$ of the relaxed problem. We obtained a new cost $\nu(t)$ at each time slot by dual gradient ascent. In Fig. \ref{fig:converge_lambda}, the cost of the server smoothly converges to a small neighborhood of the optimal cost.

Next, we verify the server cost dynamics of the proposed index-based policy when the system scalar increases. Fig. \ref{fig:converge_lambda} represents the dynamics of the cost update at scale $r=2$ and $r=20$. With the increase of the scalar $r$, the cost for the proposed index-based policy approaches is close to the optimal value of the dual cost.

\subsection{Average AoI}

We evaluate the average AoI performance of our proposed policy. We use the optimal solution to problem (\ref{originrelax}) as the lower bound of the index policy \cite{verloop_asymptotically_2016} for comparison.

We consider three benchmark policies in the experiments: \textit{Max-Age Matching Policy} (MAMP), \textit{Max-Age Reducing Policy} (MARP), which are both greedy policies, and \textit{Rounded Relax Policy} (RRP).
\begin{itemize}
    \item The MAMP chooses users with the highest current AoI.
    \item The MARP takes the transition probability of MC into consideration. Define the weight in this policy as $w_{n}=\Delta_n(t)+\frac{1}{p_n}(\Delta_n(t)-D_n(t))$, which represents the probable approximation of optimality gap reduction.
    \item The RRP is derived from the solution of the relaxed problem. 
    The RRP chooses users uniformly at random to satisfy the feasibility when violating the constraint \eqref{c1}.
\end{itemize}

Fig. \ref{aoi_converge} evaluates the average AoI under different policies such as nested index policy, MAMP, MARP, RRP, and lower bound of problem \eqref{originrelax} with $r=20$. The greedy policy MAMP and MARP is $40\%$ worse than the \textit{nested index} policy and RRP is $21\%$ worse than our approach when $r=20$. The normalized system AoI gets closer to that of the optimal AoI for the relaxed problem with the increase of the system scalar $r$. Fig. \ref{fig_sim} shows the normalized AoI of the system, i.e., the average age per user. The normalized AoI decreases almost monotonically as $r$ increases even, which supports the asymptotic optimality of our proposed policy.

\section{Conclusion}
In this study, we explored the minimization of AoI in a MEC system with heterogeneous servers and users. We formulated the problem as a two-layer MDP and introduced a novel nested index. We devised a scheduling policy that employs the nested index, ensuring the asymptotic optimality of the average expected AoI of the MEC system as the system scale expands. We also derived the computation of the nested index, which exhibits lower computational complexity. Through simulation, we demonstrated that our algorithm converges and delivers near-optimal performance.

\section*{Acknowledgments}
The research leading to these results received funding from Beijing Municipal Natural Science Foundation under Grant Agreement Grant No. 4224092. In addition, it received funding from National Key R\&D Program of China (2022ZD0116402). It was also supported by National Natural Science Foundation of China grant 62202427.

\bibliographystyle{IEEEtran}
\bibliography{Age}






%

\clearpage
\begin{appendices}
\section{}\label{ap1}
{\centering\section*{Proof of Proposition \ref{mltt}}}
We drop subscript $n$ to simplify the notation, as the proposition holds for each sub-problem. We suppose $\pi^*$ is the optimal policy under the optimal server cost $\bm{\nu}^*$. We first show the monotonicity of the optimal cost-to-go function under $\pi^*$.
\begin{lemma}
Denote the state of a user at layer $2$ as $s=(A,D)$. The cost-to-go function under $\pi^*$ is non-decreasing in $A$, i.e.,
\begin{equation}\label{monot}
\begin{aligned}
   A\le A',D=D' \Rightarrow  V((A,D),\bm{\nu})\le V((A',D),\bm{\nu}),\\
   V((A,D),\bm{\nu})-V((A-D),\bm{\nu})\\
   \le V((A',D),\bm{\nu})-V((A'-D),\bm{\nu}).
\end{aligned}
\end{equation}
\end{lemma}
The proof exploits Proposition 3.1 in \cite{jiang2015approximate}, which extends the average cost within finite steps to the infinite horizon. We show the first inequality fulfills the condition of Proposition 3.1 in \cite{jiang2015approximate}, and use backward induction to derive the second inequality.
We consider the following state transition function $f:\mathcal{S}\times\mathcal{A}\times\mathcal{W}\to\mathcal{S}$, i.e., $s_n(t+1)= f(s_n(t),\bm{y}_n(t),w_t)$, where $w_t\in\mathcal{W}$ is the information process at time $t$ \cite{jiang2015approximate}. Given $s'=(A',D')$ and $s\preccurlyeq s'$, we have:
\begin{enumerate}
    \item For every $\bm{a}\in\mathcal{A}$ and $w\in\mathcal{W}$ the state transition function satisfies $f(s,\bm{a},w)\preccurlyeq f(s',\bm{a},w)$. Since at state $s$ the next state could be either $(A+1,D)$ or $(A+1-D)$. We have $(A+1,D)\preccurlyeq (A'+1,D')$ and $A-D\le A'-D$ at the same time given $D=D'$;
    \item Denote $g_{m}(s)=C(s,\bm{a})=\nu_m+(1-p_m)A+p_m(A-D)=\nu_m+A-p_m\cdot D$ as the per stage cost. We also have $g_m(s)\le g_m(s')$;
    \item $w_{t+1}\in\mathcal{W}$ is independent of state $s\in\mathcal{S}$.
\end{enumerate}

By Proposition 3.1 in \cite{jiang2015approximate}, we conclude the inequality $V_T((A,D),\bm{\nu})\le V_T((A',D),\bm{\nu})$ of the $T$-stage cost minimization problem, where $V_T(\cdot)$ is the $T$-stage value function. By utilizing the convergence of the value iteration \cite{jiang2015approximate, optimalcontrol}, we can derive the first inequality in \eqref{monot}.

For the second inequality in \eqref{monot}, we also consider the $T$-stage value function. Given $V_{t+1}((A,D),\bm{\nu})-V_{t+1}((A-D),\bm{\nu})\le V_{t+1}((A',D),\bm{\nu})-V_{t+1}((A'-D),\bm{\nu})$. We have
\begin{equation}\label{expectedt+1}
    \begin{aligned}
        \mathbb{E}\left[V_{t+1}(f((A,D),\bm{a},w_{t+1}),\bm{\nu})\mid S_t=(A,D),a_t=\bm{a}\right]\\
        -\mathbb{E}\left[V_{t+1}(f((A-D),\bm{b},w_{t+1}),\bm{\nu})\mid S_t=(A-D),a_t=\bm{b}\right]\\
        \le \mathbb{E}\left[V_{t+1}(f((A',D),\bm{a},w_{t+1}),\bm{\nu})\mid S_t=(A',D),a_t=\bm{a}\right]\\
        -\mathbb{E}\left[V_{t+1}(f((A'-D),\bm{b},w_{t+1}),\bm{\nu})\mid S_t=(A'-D),a_t=\bm{b}\right].
    \end{aligned}
\end{equation}
We also have
\begin{equation}\label{stagecostt}
\begin{aligned}
    V_t(s,\bm{\nu})=\min\limits_{\bm{a}\in\mathcal{A}}\big[ C(s,\bm{a})\\
    +\mathbb{E}\left[V_{t+1}(f(s,\bm{a},w_{t+1}),\bm{\nu})\mid S_t=s,a_t=\bm{a}\right]\big].    
\end{aligned}
\end{equation}
Since
\begin{equation}
\begin{aligned}
    &C((A,D),\bm{a})-C((A-D),\bm{b})\\
    =&\nu_m-\nu_{m'}+(1-p_m)D\\
    =&C((A',D),\bm{a})-C((A'-D),\bm{b}),\forall \bm{a},\bm{b}\in\mathcal{A},    
\end{aligned}
\end{equation}
we can derive $V_{t}((A,D),\bm{\nu})-V_{t}((A-D),\bm{\nu})\le V_{t}((A',D),\bm{\nu})-V_{t}((A'-D),\bm{\nu})$ by adding \eqref{stagecostt} to \eqref{expectedt+1}. By using backward induction, it holds for all $t$.

Now, we will verify the MLTT property for sub-problem \eqref{originrelax}. For any user with state $s=A$ at layer 1, we compare the expected cost-to-go of choosing server $m$ and $m'$ when $p_m> p_{m'}$:
\begin{equation}
\label{19}
\begin{aligned}
       \mu_{m'}(A,\bm{\nu})- \mu_m(A,\bm{\nu})=
     \tau^{min}\cdot\nu_{m'} \\
     +(1-p_{m'})[A+\tau^{min}+V((A+\tau^{min}+1,A),\bm{\nu})]\\
     +p_{m'} V(\tau^{min},\bm{\nu})\\
     -\tau^{min}\cdot\nu_{m} \\
     -(1-p_m)[A+\tau_{n}^{min}+ V((A+\tau^{min}+1,A),\bm{\nu})]\\
     -p_{m} V(\tau^{min},\bm{\nu}).
\end{aligned}
\end{equation}
Eq. \eqref{19} is a function of variable $A$, and can be rewritten as:
\begin{equation}\label{simpmono1}
    \begin{aligned}
           \mu_{m'}(A,\bm{\nu})- \mu_m(A,\bm{\nu})=\\
           \cdots+(p_m-p_{m'})[A+ V((A+\tau^{min}+1,A),\bm{\nu})\\
           - V(\tau^{min}+1,\bm{\nu})],
    \end{aligned}
\end{equation}
where we omit terms irrelevant to $A$. Since $p_m>p_{m'}$, Eq. \eqref{simpmono1} is strictly monotonically increasing in $A$. Therefore, there must exist a threshold $H(m,m',0)=A^{m,m'}$, where we have $\mu_m(\cdot)=\mu_{m'}(\cdot)$.

For any user with state $s=(A,D)$ at layer 2, we have
\begin{equation}\label{simpmono}
    \begin{aligned}
           \mu_{m'}((A,D),\bm{\nu})- \mu_m((A,D),\bm{\nu})=\\
           \cdots+(p_m-p_{m'})[A+ V((A+1,D),\bm{\nu})\\
           - V(A-D+\tau^{min}+1,\bm{\nu})].
    \end{aligned}
\end{equation}
The difference between the two cost-to-go is also monotonically decreasing in A, hence the sub-problem \eqref{originrelax} is MLTT.

\section{}\label{ap2}
{\centering\section*{Proof of Theorem \ref{intraindex}}}
For simplicity, we specify the order of $p_m$, i.e., $p_{m-1}\le p_m,\forall 1<m\le M$. To prove the intra-indexability, we have to introduce the following lemma:
\begin{lemma}\label{upperbound}
    Given server cost $\bm{\nu}$ and state $s=(A,D)$ at layer 2, denote $\bm{\nu}'=[\nu_1,\cdots,\nu_m+\Delta,\cdots,\nu_M],\ \forall \Delta \ge 0$. The difference between two cost-to-go functions given $\bm{\nu}$ nad $\bm{\nu}'$ can be upper-bounded by
    \begin{equation}
        V_n(s,\bm{\nu}')-V_n(s,\bm{\nu})<\frac{\Delta}{p_m^2},\forall 1\le m\le M.
    \end{equation}

    Denote $p_{M+1}\triangleq 1$.Under the condition in lemma \ref{upperbound}, the difference between two cost-to-go function can be lower-bounded by
    \begin{equation}
    \begin{aligned}
         V_n(s,\bm{\nu}')-V_n(s,\bm{\nu})>-\frac{\Delta}{p_{m+1}^2},\\
         \forall, 1\le m\le M, A\ge H_n(m,m+1,D).
    \end{aligned}
    \end{equation}

\end{lemma}
The proof is shown as follows. The minimizing AoI problem can be seen as a stochastic shortest path (SSP) problem \cite{optimalcontrol}. Recall the optimal cost for user $n$ is denoted as $\gamma_n^*$. The cost-to-go function can be rewritten as:
\begin{equation}\label{sspcost}
\begin{aligned}
V_n(s,\bm{\nu})=&
\mathop{\rm{min}}\limits_\pi \mathbb{E}_{\pi}[\text{the cost from state } s \text{ to the recurrent state}\\
&\text{for the first time}]- \\
&\mathbb{E}_\pi [\text{the cost from state } s \text{ to the recurrent state}\\
&\text{with stage cost } \gamma_n^*].
\end{aligned}
\end{equation}
Unlike appendix B in \cite{zou2021minimizing}, the recurrent state for reference is not defined as the state with minimum AoI, but we can set any state at layer 1 as a reference recurrent state with zero cost-to-go. The expected cost from $s=(A,D)$ to recurrent state $\zeta$ given policy $\pi_n$ and server cost $\nu$ can be denoted as $\text{Cost}^{\pi_n}_{s\zeta}(\bm{\nu})$, and the expected step can be denoted as $N_{s\zeta}^{\pi_n}$. Considering recurrent states $\zeta=(j),\forall j\ge A-D$, we have \cite{zou2021minimizing}
\begin{equation}
\begin{aligned}
    V_n(s,\bm{\nu}')-V_n(s,\bm{\nu})    \le\text{Cost}^{\pi_n}_{s\zeta}(\bm{\nu}')-\text{Cost}^{\pi_n}_{s\zeta}(\bm{\nu}).
\end{aligned}
\end{equation}
where 
\begin{equation}
\begin{aligned}
 &\text{Cost}^{\pi_n}_{s\zeta}(\bm{\nu}')-\text{Cost}^{\pi_n}_{s\zeta}(\bm{\nu}) \\
 =&\Delta\cdot [\text{the expected time of hitting state } \Tilde{s}, \forall \pi_n(\Tilde{s})=m\\
 &\text{when transiting from s to } \zeta \text{ under policy }\pi_n].      
\end{aligned}
\end{equation}
According to Proposition \ref{mltt}, there exits $H_n(m-1,m,D),\forall D$. We can derive the upper bound by considering the following policy $\pi_n'$: (a) for all $s=(A,D), D < H_n(m-1,m,0)$ and $A>=\max\limits_d H_n(m-1,m,d)$, $\pi_n'(s)=m$ and for any other $s', \pi_n'(s')=1$. By using such a policy $\pi_n'$, we have $p_{\pi_n'(s)}<=p_{\pi_n(s)}$, i.e., the probability of finishing computation at each state gets lower, and for all $s$ and $|\{s\mid \pi_n'(s)=m\}|>|\{s\mid \pi_n(s)=m\}|$. Therefore, we have
\begin{equation}
\text{Cost}^{\pi_n}_{s\zeta}(\bm{\nu}')-\text{Cost}^{\pi_n}_{s\zeta}(\bm{\nu})\le\Delta\cdot N_{s\zeta}^{\pi_n'}.
\end{equation}
We have
\begin{equation}
\begin{aligned}
       N_{s\zeta}^{\pi_n'}&=p_m\cdot 1+\sum\limits_{i=2}^\infty p_m(1-p_m)^{i-1}(i+N_{s\zeta}^{\pi_n'}),\\
          N_{s\zeta}^{\pi_n'}&=\frac{1}{p_m^2}.
\end{aligned}
\end{equation}
Therefore,
\begin{equation}
\begin{aligned}
    V_n(s,\bm{\nu}')-V_n(s,\bm{\nu})\le\frac{\Delta}{p_m^2}.
\end{aligned}
\end{equation}
Similar to Lemma 4.4 in \cite{zou2021minimizing}, we also have
\begin{equation}
\begin{aligned}
    V_n(s,\bm{\nu}')-V_n(s,\bm{\nu})\ge(\gamma^*_n-\gamma^{*'}_n)\cdot N_{s\zeta}^{\pi_n}\ge -\frac{\Delta}{p_{m+1}^2},\\
    \forall A\ge H_n(m,m+1,D),    
\end{aligned}
\end{equation}
where $\gamma^{*'}$ is the optimal stage cost under server cost $\bm{\nu}'$.

To prove the sub-problem \eqref{originrelax} is intra-indexable, we have to show the following two claims:
\begin{itemize}
    \item If $s=(A,D)\in \mathcal{P}_{nm}^l(\bm{\nu})$, then $s \in \mathcal{P}_{nm}^l(\bm{\nu}')$ must hold for $\bm{\nu}'=[\nu_1,\cdots,\nu_m+\Delta,\cdots,\nu_M],\ \forall \Delta \ge 0 ,m\in\mathcal{M}$. 
    \item If $\nu_m\to+\infty$, then $\lim\limits_{\nu_m\to+\infty}\mathcal{P}_{nm}^l(\bm{\nu}')=\mathcal{S}_l$.
\end{itemize}


We have shown that the optimal policy for the MDP is MLTT. Recall that the threshold splits server $m-1$ and $m$ given generate age $d$ is denoted as $H_n(m-1,m,d)$. Then, the passive set for layer 2 can be expressed as \cite{zou2021minimizing}:
\begin{equation}
\begin{aligned}
     &\mathcal{P}_{nm}^l(\bm{\nu})=
    \mathop{\cup}\limits_{d=1}^\infty\big\{(a,d)\mid\\
    &a\in\{1,\cdots,H_n(m-1,m,d)-1\}\cup\{H_n(m,m+1,d),\cdots\}\big\}, \\&\forall 1<m<M,
\end{aligned}
\end{equation}

\begin{equation}
\begin{aligned}
     \mathcal{P}_{nM}^l(\bm{\nu})=
    \mathop{\cup}\limits_{d=1}^\infty\left\{(a,d)\ |\ a\in\{1,\cdots,H_n(M-1,M,d)-1\}\right\}.
\end{aligned}
\end{equation}

To show statement (i), we want to show $\mathcal{P}_{nm}^l(\bm{\nu})\subseteq\mathcal{P}_{nm}^l(\bm{\nu}')$, we can instead show that for all $d\ge 1$:

(a) $H_n(m-1,m,d) \le H_n'(m-1,m,d) ,\ \forall m \le M$;

(b) $H_n(m,m+1,d)\ge H_n'(m,m+1,d),\ \forall m\ge M$;

We first show (a). For contradiction, suppose $H_n(m-1,m,d) > H_n'(m-1,m,d)$. At state $(H_n'(m-1,m,d), d)$, taking server $m-1$ has a smaller expected cost, and we have:

\begin{equation}\label{m'}
\begin{aligned}
        (p_m-p_{m-1})[H_n'+ V_n(H_n',\bm{\nu})
    -H_n'+d\\
    -V_n((H_n'+1-d,H_n'+1-d),\bm{\nu})]\le \nu_m-\nu_{m-1},    
\end{aligned}
\end{equation}
where we denote $H_n= H_n(m-1,m,d)+1,d)$ and $H_n'= H_n'(m-1,m,d)+1,d)$ for simplicity.
If given $\bm{\nu}'$, taking server $m-1$ has a smaller expected cost, i.e.,
\begin{equation}\label{nu'}
\begin{aligned}
       (p_m-p_{m-1})[H_n' + V_n((H_n'+1,d),\bm{\nu}')-H_n'+d\\
       -V_n((H_n'+1-d,H_n'+1-d),\bm{\nu}')]\le \nu_m'-\nu_{m-1}.    
\end{aligned}
\end{equation}

Let Eq. \eqref{nu'} - Eq. \eqref{m'}, we have
\begin{equation}
\begin{aligned}
      V_n((H_n'+1,d),\bm{\nu}')
      - V_n((H_n'+1,d),\bm{\nu})\\
      -[V_n((H_n'+1-d,H_n'+1-d),\bm{\nu}')\\
      -V_n((H_n'+1-d,H_n'+1-d),\bm{\nu})]\\
      \ge  \frac{\Delta}{p_m-p_{m-1}},
\end{aligned}
\end{equation}
when $p_m-p_{m-1}\le p_m^2$, it contradicts the upper bound in Lemma 3.

Then, we will show the sufficiency of condition (b). Suppose $H_n(m-1,m,d)\le H_n'(m-1,m,d)$. At state $s=(H_n'(m-1,m,d)-1,d)$, policy $\pi_n$ prefer a server $m-1>m$ over server $m$, i.e., $H_n'(m-1,m,d)\ge H_n(m-1,m,d)\ge H_n(m,m+1,d)$, and we have
\begin{equation}
\begin{aligned}
        (p_{m+1}-p_{m})[H_n'+ V_n((H_n' +1,d),\bm{\nu})]
        \le \nu_{m+1}-\nu_{m}.
\end{aligned}
\end{equation}

Similarly, policy $\pi_n'$ prefers server $m$ at state $s=(H_n'(m,m+1,d)-1,d)$, i.e.,

\begin{equation}
\begin{aligned}
        (p_{m+1}-p_{m})[H_n' + V_n((H_n'+1,d),\bm{\nu}')]\\
     \le \nu_{m+1}'-\nu_{m}'=-\Delta.
\end{aligned}
\end{equation}

For $p_{m+1}-p_{m}>0$, we have
\begin{equation}
\begin{aligned}
     V_n((H_n'+1,d),\bm{\nu}')
     - V_n((H_n'+1,d),\bm{\nu})\\
     \le -\frac{\Delta}{p_{m+1}-p_m},
    \end{aligned}
\end{equation}
when $p_{m+1}-p_m\le p_{m+1}^2$, contradicts the lower bound derived in lemma 4.
Therefore, the cardinality of passive set $\mathcal{P}_{nm}(\bm{\nu})$ grow monotonically to $|\mathcal{S}|$ as server cost $\nu_m$ increases from $0$ to $+\infty$.
Here we consider a strict condition that $p_m-p_{m-1}\le p_m^2,\forall 1<m\le M$ due to that we obtain a rough upper bound on $N_{s\zeta}^{\pi_n'}$. In future studies, we will seek a tighter bound on $N_{s\zeta}^{\pi_n'}$.

\section{}\label{ap3} 
{\centering\section*{Proof of Proposition \ref{indexfunc}}}

Appendix \ref{ap1} shows that the problem \eqref{originrelax} satisfies the MLTT property. For simplicity, we denote the minimal age to offload tasks to server m for a user at layer 1 as $H^*_{m}$ and the optimal age to offload tasks to server m for a user at layer 2 given generated age $d$ as $H^*_{m}(d)$, and the minimum computational time is 1. 
Let the cost-to-go for the recurrent state $s=(1)$ to be 0, i.e., . We first focus on the cost-to-go for one state at layer 1 with age $H^*_{m-1}\le A\le H^*_m$:
\begin{equation}\label{init}
\begin{aligned}
        V(A,\bm{\nu})=&A-\gamma^*+\nu_{m-1}+p_{m-1}\cdot 0\\
        &+(1-p_{m-1})V((A+1,A),\bm{\nu}).
\end{aligned}
\end{equation}
For $A^*_M\le A$, we assert that the optimal server for state $s=(A+1,A)$ is also M, and we have
\begin{equation}
\begin{aligned}
    V(A,\bm{\nu})=&A-\gamma^*+\nu_{M}\\
        &+(1-p_{M})V((A+1,A),\bm{\nu})\\
        =&\frac{A-\gamma^*+\nu_{M}}{p_M}+\frac{1-p_M}{p^2_M}.
\end{aligned}
\end{equation}
\
For $H^*_{m-1}\le A\le H^*_m$, we have
\begin{equation}
  \begin{aligned}
     V&(A,\bm{\nu})=\sum\limits_{i=1}^{A^*_{m}(A)-A*_{m-i}}(1-p_{m-1})^{i-1}\\
     &\cdot (A*_{m-1}+i-1+p_{m-1}V(i,\bm{\nu})+\nu_{m-1}-\gamma^*)\\
     &+(1-p_{m-1})^{A^*_{m}(A)-1}V((A^*_{m}(A),A),\bm{\nu}),
\end{aligned}  
\end{equation}
and we have
\begin{equation}
    \begin{aligned}
        V&((A^*_{m}(A),A),\bm{\nu})=\sum\limits_{i=1}^{A^*_{m+1}(A)-A^*_{m}(A)}(1-p_{m})^{i-1}\\
     &\cdot(A^*_{m}(A)+i-1+p_{m}V(i,\bm{\nu})+\nu_{m}-\gamma^*)\\
     +&(1-p_{m})^{A^*_{m+1}(A)-A^*_{m}(A)-1}V((A^*_{m+1}(A),A),\bm{\nu}),
    \end{aligned}
\end{equation}
\begin{equation}
    \begin{aligned}
        V((A^*_{M}(A),A),\bm{\nu})=\frac{A^*_{M}(A)-\gamma^*+\nu_{M}}{p_M}+\frac{1-p_M}{p^2_M}.
    \end{aligned}
\end{equation}
For $A< H^*_1$, we have
\begin{equation}
    \begin{aligned}
        V(A,\bm{\nu})=A+V(A+1,\bm{\nu})-\gamma^*.
    \end{aligned}
\end{equation}
\begin{equation}\label{end}
    V(1,\bm{\nu})=0.
\end{equation}
Since $p_1,p_2,\dots,p_M$ and $\bm{\nu}$ is given, we can derive the closed-form relationship of $\gamma^*$ with $H^*_m,H^*_m(d),\forall m\in\mathcal{M},d\le A^*_M$ from Eq. \eqref{init} to Eq. \eqref{end}.
The \textit{optimal threshold} follows:
\begin{equation}
\begin{aligned}
     \nu_{m-1}+V((A,D),\bm{\nu})&\le \nu_m+ V((A,D-1),\bm{\nu})\\
    \nu_m+ V((A,D-1),\bm{\nu})&\le \nu_m+ V((A+1,D),\bm{\nu})\\
    \nu_m+ V((A+1,D),\bm{\nu})&\le \nu_{m-1}+ V((A+1,D+1),\bm{\nu}).\\
\end{aligned}
\end{equation}
Subtracting $\nu_m+V((A,D-1),\bm{\nu})-\gamma^*$ to every term above:
\begin{equation}
\label{optimal}
    \nu_{m-1}-\nu_{m}+A\le\gamma^*\le \nu_{m-1}-\nu_{m}+A+1.
\end{equation}
Let $\nu_{m-1}-\nu_{m}+A=\gamma^*$, we derive the expression of $\nu_m$, which is the index function \cite{tripathi_whittle_2019}.

\section{}\label{ap4}
{\centering\section*{Proof of Proposition \ref{fixedpoint}}}

In Appendix \ref{ap2}, we have shown that the subproblem \eqref{originrelax} satisfies the \textit{intra-indexability}. The asymptotic optimum will hold for the original problem under the nested index policy if \textit{precise division} property satisfies \cite{zou2021minimizing}. The \textit{precise division} property is defined as follows:
\begin{definition}[Precise Division]
    Given state $s$ and the server cost $\bm{\nu}$, suppose the sub-problem is intra-indexable. We say the preference for server $m$ is precisely divisible at layer $l$ by the nested-index $I_{nm}(s,\bm{\nu},l)$ if the following holds:

\textit{(i) If $I_{nm}(s,\bm{\nu})=\nu_m$, then $\mu_{nm}(s,\bm{\nu})\le\mu_{nm'}(s,\bm{\nu})$.}

\textit{(ii) If $I_{nm}(s,\bm{\nu})>\nu_m$, then $\mu_{nm}(s,\bm{\nu})<\mu_{nm'}(s,\bm{\nu})$.}

\textit{(iii) Otherwise, there exists $m'\ne m$ s.t. $\mu_{nm}(s,\bm{\nu})>\mu_{nm'}(s,\bm{\nu})$.}

\end{definition}


The precise division property established the connection between the index value and the optimal policy. 
Then, we will show that the sub-problem \eqref{originrelax} satisfies the \textit{precise division} property.

\begin{proposition}
Given state $\mathcal{S}$ and server cost $\bm{\nu}$, the sub-problem \eqref{originrelax} satisfies the precise division property defined in Definition 7.
\end{proposition}

The precise division property is more strict than intra-indexability, due to the transition of optimal action happens when the index value coincides with the server cost. The author in \cite{zou2021minimizing} proves this proposition by showing the difference $V_n(s,\bm{\nu}')-V_n(s,\bm{\nu})$ is uniformly bounded:
\begin{lemma}
    Denote $p_{M+1}\triangleq 1$.Under the condition in lemma \ref{upperbound}, the difference between two cost-to-go function can be lower-bounded by
    \begin{equation}
         V_n(s,\bm{\nu}')-V_n(s,\bm{\nu})>-\frac{\Delta}{p_{m+1}},\forall, 1\le m\le M.
    \end{equation}
\end{lemma}
According to Case 3, Lemma 4.6 in \cite{zou2021minimizing}, we have to show the monotonicity of $h(\pi_n',s=(\Delta,D),\bm{\nu}')$ in $\Delta$, where
\begin{equation}
    h(\pi_n',s=(\Delta,D),\bm{\nu}')=\mu_{\pi_n'(s)}(s,\bm{\nu}')-\mu_m(s,\bm{\nu}')+\Delta-{\gamma_n^*}'+\gamma_n^*.
\end{equation}
We have
\begin{equation}\begin{aligned}
    \mu_{\pi_n'(s)}(s,\bm{\nu}')-\mu_m(s,\bm{\nu}') =\nu_{\pi_n'(s)}-\nu_m\\
    +(p_m-p_{\pi_n'(s)})(D+V((A+1,D),\bm{\nu}')-V((A+1-D),\bm{\nu}')),
\end{aligned}
\end{equation}
where we have shown $V((A+1,D),\bm{\nu}')-V((A+1-D),\bm{\nu}')$ is monotonically increasing with $A$ in Appendix \ref{ap1}.

\end{appendices}
\clearpage

\end{document}